\documentclass{article}

\PassOptionsToPackage{numbers}{natbib}

\usepackage[preprint]{neurips_2025}

\usepackage[utf8]{inputenc} 
\usepackage[T1]{fontenc}    
\usepackage{hyperref}       
\usepackage{url}            
\usepackage{amsfonts}       
\usepackage{nicefrac}       
\usepackage{microtype}      
\usepackage{amsmath,amssymb,amsfonts,amsthm}
\usepackage{algorithm,algorithmic}
\usepackage{microtype}
\usepackage{graphicx}
\usepackage{multirow}
\usepackage{url}
\usepackage{caption}
\usepackage{subcaption}
\usepackage{cleveref}

\usepackage[table]{xcolor}  
\usepackage{pifont}     
\usepackage{array}
\usepackage{makecell}
\usepackage{arydshln}
\usepackage{adjustbox}
\usepackage{rotating}
\usepackage{wrapfig}
\usepackage{booktabs}

\crefname{section}{Sec.}{Secs.}
\Crefname{section}{Section}{Sections}
\Crefname{table}{Table}{Tables}
\crefname{table}{Tab.}{Tabs.}
\crefname{figure}{Fig.}{Figs.}
\Crefname{figure}{Figure}{Figures}
\crefname{appendix}{Appx.}{Appxs.}
\Crefname{appendix}{Appendix}{Appendixs}
\crefname{subsection}{Sec.}{Secs.}

\newcommand{\tablestyle}[2]{\setlength{\tabcolsep}{#1}\renewcommand{\arraystretch}{#2}\centering\footnotesize}
\newlength\savewidth\newcommand\shline{\noalign{\global\savewidth\arrayrulewidth
		\global\arrayrulewidth .8pt}\hline\noalign{\global\arrayrulewidth\savewidth}}

\newcolumntype{a}{>{\columncolor{gray!10}}c}
\newcolumntype{b}{>{\columncolor{gray!25}}c}

\definecolor{maroon}{cmyk}{0,0.1,0.01,0.01}
\definecolor{blue}{cmyk}{0.95,0.0,0.2,0.2}
\definecolor{yellow}{cmyk}{0.01,0.0,0.2,0.01}
\definecolor{lightblue}{cmyk}{0.1,0.0,0.02,0.02}
\definecolor{case_verb}{HTML}{fbde84}
\definecolor{case_adj}{HTML}{cccdff}
\definecolor{case_noun}{HTML}{bfeaf1}
\definecolor{case_ff}{HTML}{e65352}
\definecolor{case_error}{HTML}{ffff00}
\definecolor{darkgreen}{RGB}{51,181,41}
\definecolor{darkorange}{RGB}{252,135,62}
\definecolor{t_green}{HTML}{f1f2e4}

\definecolor{LIGHT_BLUE}{HTML}{cce4fe}
\definecolor{LIGHT_RED}{HTML}{f1b9b8}
\definecolor{LIGHT_YELLOW}{HTML}{f1f58a}
\definecolor{LIGHT_GREEN}{HTML}{f1f2e4}
\definecolor{LIGHT_PURPLE}{HTML}{b6a7b9}

\definecolor{lightgray}{gray}{0.95}

\title{CAVALRY-V: A Large-Scale Generator Framework for Adversarial Attacks on Video MLLMs}

\author{\textbf{Jiaming Zhang\textsuperscript{\rm 1} \quad Rui Hu\textsuperscript{\rm 2} \quad Qing Guo\textsuperscript{\rm 3}\thanks{Corresponding authors} \quad Wei Yang Bryan Lim\textsuperscript{\rm 1}\footnotemark[1]} \\
\textsuperscript{\rm 1}Nanyang Technological University \
\textsuperscript{\rm 2}Beijing Jiaotong University \ \textsuperscript{\rm 3}A*STAR \\
{\centering Project Page: \href{https://jiamingzhang94.github.io/cavalry/}{\textbf{CAVALRY-V}}}
}

\begin{document}

\maketitle

\begin{abstract}
Video Multimodal Large Language Models (V-MLLMs) have shown impressive capabilities in temporal reasoning and cross-modal understanding, yet their vulnerability to adversarial attacks remains underexplored due to unique challenges: complex cross-modal reasoning mechanisms, temporal dependencies, and computational constraints. We present \textbf{CAVALRY-V} (\textbf{C}ross-mod\textbf{A}l \textbf{L}anguage-\textbf{V}ision \textbf{A}dve\textbf{R}sarial \textbf{Y}ielding for \textbf{V}ideos), a novel framework that directly targets the critical interface between visual perception and language generation in V-MLLMs. Our approach introduces two key innovations: (1) a dual-objective semantic-visual loss function that simultaneously disrupts the model's text generation logits and visual representations to undermine cross-modal integration, and (2) a computationally efficient two-stage generator framework that combines large-scale pre-training for cross-model transferability with specialized fine-tuning for spatiotemporal coherence. 
Empirical evaluation on comprehensive video understanding benchmarks demonstrates that CAVALRY-V significantly outperforms existing attack methods, achieving \textbf{22.8\%} average improvement over the best baseline attacks on both commercial systems (GPT-4.1, Gemini 2.0) and open-source models (QwenVL-2.5, InternVL-2.5, Llava-Video, Aria, MiniCPM-o-2.6). Our framework achieves flexibility through implicit temporal coherence modeling rather than explicit regularization, enabling significant performance improvements even on image understanding (\textbf{34.4\%} average gain). This capability demonstrates CAVALRY-V's potential as a foundational approach for adversarial research across multimodal systems.

\end{abstract}

\section{Introduction}

The recent advances in foundation models that integrate vision and language capabilities have driven significant progress in artificial intelligence research. Video Multimodal Large Language Models (V-MLLMs) have shown impressive abilities in temporal reasoning and cross-modal understanding \cite{radford2021learning, achiam2023gpt, team2024gemini, chen2024expanding, zhang2024video, li2024aria, abdin2024phi, lin2024vila, bai2025qwen2, li2024llavaonevision, zhang2025videollama}. As these systems begin to be deployed in critical domains such as autonomous driving, healthcare, and security, designing adversarial attacks to evaluate their robustness becomes increasingly important. While adversarial examples have been extensively studied \cite{goodfellow2014explaining, madry2017towards}, the vulnerability landscape of V-MLLMs remains underexplored. This research gap stems from several technical challenges: the complexity of cross-modal reasoning mechanisms, the necessity to model temporal dependencies, and the computational constraints for long videos, all further complicated by the requirement for transferability in black-box attack scenarios.

Existing attack methodologies fall short when facing these emerging challenges. Traditional video attack techniques primarily target unimodal classification models and fail to capture how adversarial perturbations disrupt the complex interaction between visual and language understanding mechanism \cite{wei2020heuristic,wei2022cross, chen2024rethinking, wei2022boosting, wei2022adaptive,wei2023adaptive,chen2023gcma,gao2024retome}. Similarly, existing multimodal attacks developed for Vision-Language Models (VLMs) mainly focus on static images, neglecting the temporal dependencies critical for video comprehension \cite{zhang2022towards, zhao2023evaluating, lu2023set, yin2023vlattack, zhang2025anyattack, huang2025xtransfer}. Furthermore, the computational demands for attacking V-MLLMs (which must process videos potentially spanning thousands of frames) render iterative optimization (PGD-like) \cite{madry2017towards} approaches impractical for real-world applications.

To addresses the aforementioned challenges, we present \textbf{CAVALRY-V} (\textbf{C}ross-mod\textbf{A}l \textbf{L}anguage-\textbf{V}ision \textbf{A}dve\textbf{R}sarial \textbf{Y}ielding for \textbf{V}ideos), a novel framework for generating adversarial videos in attacking V-MLLMs. CAVALRY-V tackles the complexity of cross-modal reasoning, accounts for temporal dynamics, and overcomes computational constraints while ensuring transferability across model architectures. Our approach advances the state-of-the-art through two primary innovations:

First, we introduce a paradigm shift by directly targeting language generation mechanisms instead of the classification boundaries typically exploited in unimodal vision models. Given a query, our method employs a semantic loss function in the response space that maximizes divergence between clean and perturbed outputs. The implementation is built upon two complementary objectives: (1) semantic loss targeting text generation logits to disrupt reasoning processes, and (2) visual loss targeting the V-MLLM's visual encoder alongside an auxiliary smoother feature extractor. This dual manipulation of both cross-modal integration and visual encoding systematically undermines the model's ability to ground language in visual evidence, the core capability of V-MLLMs.

Second, we address computational efficiency through training a generator for scaling video attacks. In the first pre-training stage, we leverage the large-scale LAION-400M dataset \cite{schuhmann2021laion} to train a foundation generator via our aforementioned objective function. This generator produces diverse perturbation patterns aligned with the knowledge distribution of potential target V-MLLMs—a crucial factor for enhancing transferability. The subsequent stage comprises a two-phase fine-tuning process. We first optimize the generator on visual instruction tuning datasets containing multiple question-answering dialogues. Next, we fine-tune on video datasets, where the generator learns to process temporally correlated frame sequences and produce coherent spatiotemporal perturbations. This framework indirectly enables the generator to learn spatiotemporal correlations without introducing additional modules, providing flexibility that allows effective operation even on single-frame input (i.e., image) while achieving remarkable performance.

Our empirical evaluation encompasses seven state-of-the-art V-MLLMs, including commercial systems (GPT-4.1 \cite{achiam2023gpt} and Gemini 2.0 \cite{team2024gemini}) and open-source models (QwenVL-2.5 \cite{bai2025qwen2}, InternVL-2.5 \cite{chen2024expanding}, Llava-Video \cite{zhang2024video}, Aria \cite{li2024aria} and MiniCPM-o-2.6 \cite{yao2024minicpm}). Unlike previous adversarial research that relies on simple accuracy metrics for limited tasks, we employ MMBench-Video \cite{fang2024mmbench} benchmark, which assesses diverse video understanding abilities across both long and short videos. Finally, to demonstrate the flexibility of our framework, we evaluate on the MME benchmark \cite{fu2023mme} for static image understanding. Superior performance on both video (22.8\% average improvement) and image (34.4\% average improvement) understanding benchmarks demonstrates the effectiveness of our proposed approach. Our flexible framework, along with the publicly released model weights, facilitates effortless deployment across diverse models and evaluation scenarios, providing insights for future V-MLLM robustness research. Our work makes three primary contributions:
\begin{itemize}
\item We introduce a dual-objective semantic-visual loss function that directly disrupts the cross-modal integration between visual perception and linguistic reasoning in V-MLLMs.
\item We develop a computationally efficient two-stage generator framework that combines large-scale training for cross-model transferability with visual-linguistic and temporal fine-tuning to flexibly produce spatiotemporally coherent adversarial perturbations.
\item We demonstrate the effectiveness across seven state-of-the-art V-MLLMs, including five open-source models and two commercial systems, using a comprehensive video understanding and an image understanding benchmark.
\end{itemize}

\section{Related Work}
\label{sec:related_work}

\paragraph{Video Multimodal Large Language Models}
The field of V-MLLMs evolved from foundational vision-language models like CLIP \cite{radford2021learning}, which established crucial cross-modal alignment paradigms essential for temporal understanding in videos.
Academic research has driven significant innovations through specialized architectures: InternVL \cite{chen2024expanding} and Qwen-VL series \cite{bai2025qwen2} pioneered efficient frame processing strategies, while LLaVA-Video \cite{zhang2024video} advanced temporal reasoning through multi-stage training.
Open-source efforts like Phi-3.5 \cite{abdin2024phi}, VILA \cite{lin2024vila}, MiniCPM-o \cite{yao2024minicpm} and Aria \cite{li2024aria} have demonstrated the effectiveness of extensive video-instruction tuning and novel temporal aggregation methods.
Commercial models such as GPT-4 \cite{achiam2023gpt} and Gemini \cite{team2024gemini} represent the current state-of-the-art with their end-to-end architectures integrating perception and reasoning within unified generative frameworks.

\paragraph{Adversarial Attacks on Videos and Vision-Language Models}
Research on adversarial attacks has made significant progress along two complementary directions, each providing valuable insights for our work on V-MLLMs.
The first direction has explored video (pure-vision) model robustness through temporal modeling attacks that extend image-based adversarial techniques to video sequences on action recognition and classification tasks \cite{wei2020heuristic,wei2022cross, chen2024rethinking, wei2022boosting, wei2022adaptive,wei2023adaptive,chen2023gcma,gao2024retome}.
The second direction has explored adversarial vulnerabilities in VLMs, with recent work developing effective methods for disrupting image-text alignment in multimodal systems. 
These methods revealed important insights about the vulnerability of cross-modal integration mechanisms \cite{zhang2022towards, zhao2023evaluating, lu2023set, yin2023vlattack, zhang2025anyattack, huang2025xtransfer}.
However, with the rapid advancement of V-MLLMs, these methods face significant challenges in effectively handling video data. Among these methods, generator-based approaches offer promising directions by enabling efficient adversarial example generation without requiring iterative PGD-like optimization for each frame.
As a complementary approach, Universal Adversarial Perturbations (UAPs) \cite{moosavi2017universal}, particularly those designed for VLMs \cite{zhou2023advclip, fang2024one, huang2025xtransfer}, present another promising solution due to their minimal computational overhead and applicability to video data.
Building upon these valuable contributions, our CAVALRY-V framework addresses the unique challenges at their intersection: simultaneously modeling cross-modal reasoning vulnerabilities and temporal dependencies while maintaining computational feasibility and black-box transferability for real-world video content.

\section{Problem Formulation}

Let $V = \{x_1, x_2, ..., x_{N}\}$ represent a video with $N$ frames, where each frame $x_i \in \mathbb{R}^{C \times H \times W}$ is an image with $C$ channels and dimensions $H \times W$. In practice, a V-MLLM typically samples a subset of frames $S = \{x_{t_1}, x_{t_2}, ..., x_{t_k}\}$ where $k \ll N$ due to computational constraints. Notably, the frame sampling strategy $\phi: V \rightarrow S$ varies significantly across different V-MLLM architectures, with models employing uniform sampling, scene-change detection, or learned importance metrics. A target V-MLLM, denoted as $\mathcal{T}$, processes these sampled frames along with a question $Q$ to generate a textual answer $A$:
\begin{equation}
    A = \mathcal{T}(\phi(V), Q).
\end{equation}
Unlike traditional classification models that output probability distributions over discrete labels, V-MLLMs generate natural language responses that interpret visual content through complex cross-modal reasoning processes. This architectural complexity, combined with the heterogeneity in frame selection, presents significant challenges for adversarial attacks. Perturbing only specific frames may prove ineffective against models employing different sampling strategies, while the open-ended nature of text generation complicates the definition of attack success.

\paragraph{Attack Objective}
Given a video $V$, a question $Q$, and a ground-truth answer $A_{gt}$, our goal is to generate a perturbed video $V' = V + \delta = \{x'_1, x'_2, ..., x'_{N}\}$ on a white-box surrogate model $\mathcal{M}$, where $\delta = \{\delta_1, \delta_2, ..., \delta_{N}\}$ represents frame-wise perturbations. 
The attack aims to optimize the following objective:
\begin{equation}
   \delta^* = \arg\max_{\delta} D(\mathcal{M}(\phi(V+\delta), Q), A_{gt}) \quad \text{s.t.}\quad\|\delta_i\|_\infty\le\epsilon,
\end{equation}
where $D(\cdot,\cdot)$ is a semantic divergence function measuring the dissimilarity between the model's response to the perturbed video and the ground-truth answer. The perturbation is constrained to be visually imperceptible through the $\ell_\infty$-norm bound $\epsilon$.


\paragraph{Threat Model} 
We focus on the challenging black-box transfer attack scenario, which represents the most realistic threat for deployed V-MLLMs. In this setting, the attacker generates perturbations using a white-box surrogate model $\mathcal{M}$ and evaluates their transferability to two types of black-box targets $\mathcal{T}$: (1) Open-source models deployed locally using official weights (e.g., Qwen-2.5, InternVL-2.5); (2) Commercial models accessible solely through their APIs (e.g., GPT-4.1, Gemini 2.0). Rather than using simplistic metrics like answer flipping, we measure attack success through performance degradation on standardized video understanding benchmarks using their native evaluation metrics. 

\section{Methodology}

\begin{figure*}[t]
\centering
\includegraphics[width=\textwidth]{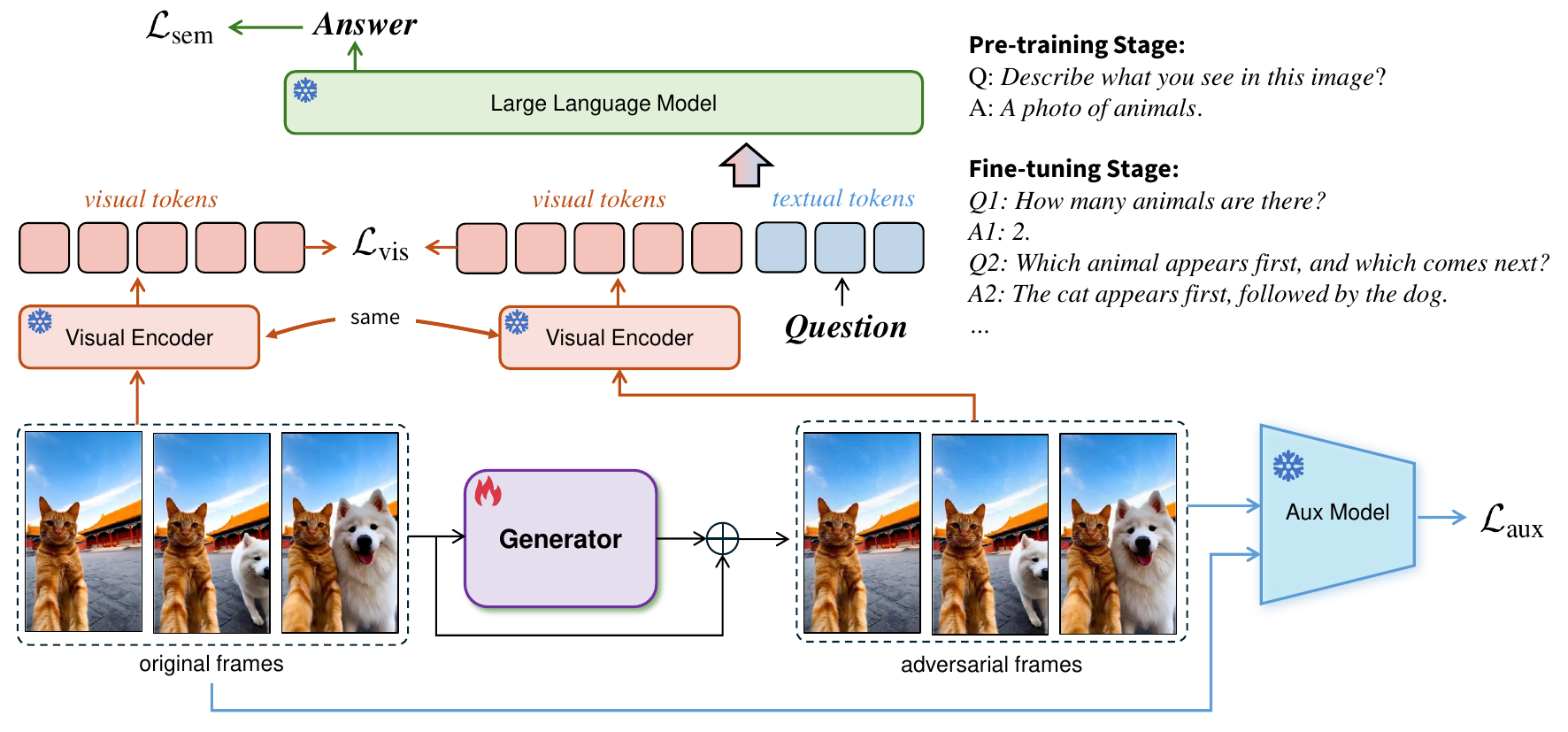} 
\caption{\textbf{Overview of the CAVALRY-V framework.} The generator $G$ produces perturbation patterns of identical dimensions to the input frames, and constitutes the only trainable component within the system, with all other network parameters remaining frozen. The colored arrows sequentially indicate the computation flow for three different objective functions. The training process is divided into two principal stages: (1) pre-training on a large-scale dataset with fixed \texttt{Question} templates to establish foundational adversarial patterns, and (2) fine-tuning with more diverse question-answer pairs and temporally correlated video data to enhance spatiotemporal coherence and attack transferability.} 
\label{fig:framework} 
\end{figure*}

\paragraph{Framework Overview}
We introduce CAVALRY-V, a transferable adversarial attack framework designed to address the unique challenges posed by V-MLLMs. Figure \ref{fig:framework} illustrates our approach. At its core, CAVALRY-V employs a UNet-style \cite{ronneberger2015u} generator $G$ that transforms an input video $V \in \mathbb{R}^{N \times C \times H \times W}$ into restricted adversarial perturbations $\delta \in \mathbb{R}^{N \times C \times H \times W}$, constrained such that $\|\delta\|_{\infty} \leq \epsilon$ to ensure imperceptibility.



\subsection{Disrupting Cross-Modal Integration}
Traditional adversarial attacks for image classification models typically target class boundaries or feature representations within a single modality. However, the power of V-MLLMs lies in their ability to integrate visual perception with language understanding, a fundamentally different mechanism that requires a novel attack approach. Rather than manipulating classification logits, we target the critical pathway through which visual information influences language generation, effectively severing the connection between what the model ``sees'' and what it ``says.''

\paragraph{Cross-Modal Disruption via Semantic Loss}
Let $\mathcal{M}_{\text{LM}}$ represent the language generation component of a surrogate V-MLLM and $\mathcal{M}_{\text{VE}}$ denote its visual encoding component. For a given video $V$, question $Q$, and ground-truth answer $A_{gt}$, we formulate our semantic loss to directly disrupt the language generation by maximizing the deviation from expected outputs:
\begin{equation}
\mathcal{L}_{\text{sem}}(\theta) = -\mathbb{E}_{t} \left[ \log P_{\mathcal{M}_{\text{LM}}}\left(a_t | \mathcal{M}_{\text{VE}}(V + G_\theta(V)), Q, a_{<t}\right) \right],
\end{equation}
where $a_t$ represents tokens in the ground-truth answer and $a_{<t}$ represents previously generated tokens. We maximize the semantic divergence between outputs from perturbed inputs and ground-truth responses. This directly compromises the model's ability to generate contextually appropriate answers based on visual evidence.

\paragraph{Visual Representation Manipulation}
To amplify the attack effectiveness, we simultaneously target the visual representation space through a feature-level loss:
\begin{equation}
\mathcal{L}_{\text{vis}}(\theta) = \left\|\mathcal{M}_{\text{VE}}(V) - \mathcal{M}_{\text{VE}}(V + G_\theta(V))\right\|_2^2.
\end{equation}
This component maximizes the distance between clean and perturbed visual features, creating significant shifts in the visual representations while remaining imperceptible in the pixel space. While the semantic loss focuses on end-to-end output manipulation, this complementary approach directly targets intermediate visual encodings that serve as critical inputs to the cross-modal integration process. By perturbing these representations, we corrupt the visual evidence before it even reaches the reasoning components, creating a two-pronged attack that is particularly effective against the hierarchical processing pipeline of V-MLLMs.
To further ensure cross-architecture transferability, we incorporate an auxiliary feature loss using an adversarially trained auxiliary model $\mathcal{F}$:
\begin{equation}
\mathcal{L}_{\text{aux}}(\theta) = \left\|\mathcal{F}(V) - \mathcal{F}(V + G_\theta(V))\right\|_2^2
\end{equation}
This auxiliary component leverages an adversarially trained model with smoother feature spaces. This property is helpful for transferability across different model architectures as it has improved Lipschitz properties \cite{cisse2017parseval}.

\paragraph{Overall Objective}
The complete loss function, which we aim to maximize, combines these components:
\begin{equation}
\label{eq:total}
\mathcal{L}(\theta) = \lambda_1\mathcal{L}_{\text{sem}}(\theta) + \lambda_2\mathcal{L}_{\text{vis}}(\theta) + \lambda_3\mathcal{L}_{\text{aux}}(\theta)
\end{equation}
where $\lambda_1, \lambda_2, \lambda_3$ are hyperparameters balancing the relative importance of each loss component.
This multi-faceted approach provides several key advantages. By simultaneously targeting both language generation and visual representation spaces, our method creates cascading failures throughout the cross-modal reasoning pipeline. Ultimately, CAVALRY-V exploits the fundamental vulnerability of V-MLLMs: their reliance on accurate visual grounding for language generation, a critical weakness not present in unimodal systems.

\subsection{Progressive Two-stage Generator Training}
Directly optimizing perturbations for videos is computationally prohibitive due to the temporal dimension and the complexity of V-MLLMs. To address these challenges, we propose a two-stage training approach that progressively adapts the generator from large-scale image collections to video dynamics.

\paragraph{Large-scale Pre-training}
We first pre-train our generator on the LAION-400M dataset \cite{schuhmann2021laion}, which contains a vast collection of image-caption pairs spanning diverse visual concepts and domains. During this stage, we treat each image as a single-frame input, construct a uniform question $Q$ = ``Describe what you see in this image'' for all samples, and use the image caption as the ground-truth answer $A_{gt}$. We optimize using the $\mathcal{L}(\theta)$ defined in Eqn. \ref{eq:total}. 
This large-scale pre-training phase is critical as it enables the generator to learn diverse perturbation patterns that align with the knowledge distribution of potential target V-MLLMs, establishing a strong foundation for cross-architecture transferability. While our formulation uses $V$ to represent video inputs, the generator actually processes individual frames as images during both training and inference.

\paragraph{Fine-tuning for Visual-Linguistic and Temporal Coherence}
The foundation generator is further refined through the fine-tuning process. First, we fine-tune on the LLaVA-Instruct-150K dataset \cite{liu2023visual}, which provides diverse question-answering dialogues covering a wide range of visual reasoning tasks. We continue treating images as single-frame inputs while leveraging the dataset's diverse question types $Q$ and corresponding answers $A_{gt}$ via Eqn. \ref{eq:total}. Finally, we fine-tune on the Video-MME dataset \cite{fu2024videomme} using the same objective function. The key distinction in this phase is our training approach: each batch contains frames sampled from the same video, sharing identical questions and ground-truth answers. This structured batching strategy implicitly encourages the generator to produce temporally coherent perturbations without requiring explicit temporal regularization terms. By seeing multiple frames from the same video context during training, the generator learns to produce consistent perturbations across a temporal sequence.

\section{Experiments}
This section presents comprehensive experiments evaluating CAVALRY-V on both open-source and commercial V-MLLMs. Our evaluation addresses three key questions: (1) the effectiveness of our approach in disrupting video understanding compared to existing attack methods, (2) further analyses about ablation studies, mechanisms and temporal consistency, and (3) the generalizability of our approach to static image benchmark.

\subsection{Experimental Setup}
\label{sec:experiment_setup}
\paragraph{Datasets and Models} We selected a diverse set of state-of-the-art V-MLLMs for evaluation based on the OpenVLM Video Leaderboard\footnote{\url{https://huggingface.co/spaces/opencompass/openvlm_video_leaderboard}}. Our evaluation included five leading open-source models (Qwen-2.5, InternVL-2.5, Llava-Video, Aria, MiniCPM-o-2.6) and two commercial systems (GPT-4.1, Gemini 2.0). Due to computational constraints, we utilized smaller parameter variants (7B or 8B) for some models, although their larger counterparts (up to 72B) rank higher on the leaderboard. For our framework, we employed LAION-400M dataset \cite{schuhmann2021laion} for foundation pre-training, followed by LLaVA-Instruct-150K dataset \cite{liu2023visual} and Video-MME dataset \cite{fu2024videomme} for fine-tuning. Our primary evaluation metric was performance on MMBench-Video \cite{fang2024mmbench}, a comprehensive video understanding benchmark with durations ranging from 30 seconds to 6 minutes. We additionally employed the image MME benchmark \cite{fu2023mme} to validate our framework's adaptability to single-frame (i.e., image) scenarios. We used the VLMEvalKit toolkit \footnote{\url{https://github.com/open-compass/VLMEvalKit}} to conduct evaluations on both benchmarks. LLMs served as automatic evaluators to assess the semantic alignment between model responses and ground-truth answers. While MME defaults to GPT-4o-mini and MMBench-Video uses GPT-4-turbo, we standardized our evaluation process by using GPT-4o-mini for both benchmarks, maintaining default parameters such as temperature=0.


\paragraph{Baseline Methods}
Based on our analysis of attack categories, we selected state-of-the-art approaches as baselines. For single-modality models, we employed GCMA \cite{chen2023gcma} (ACM MM 2023), an efficient generator-based method. We also included CWA \cite{chen2024rethinking} (ICLR 2024), an image attack extensible to video domains; however, due to its iterative generation process and computational demands, we applied identical perturbations to all frames within each second. For multimodal attacks, we implemented AnyAttack \cite{zhang2025anyattack} (CVPR 2025). Since their released generator weights were trained for targeted attacks, we deployed it with randomized targets to align with our untargeted misclassification objective. Additionally, we incorporated X-Transfer \cite{huang2025xtransfer} (ICML 2025), a UAP approach for multimodal models, applying their officially released perturbation pattern uniformly across all video frames.

\paragraph{Implementation Details}
In this work, we employed InternVL-2.5-1B as our surrogate model $\mathcal{M}$ and evaluated black-box transferability to other target V-MLLMs, where $\epsilon=16/255$. First, we trained generator $G$ for 1,400,000 steps on the LAION-400M dataset, using a batch size of 8 per GPU across 8 NVIDIA A800 80GB GPUs, with the AdamW optimizer and a learning rate of 0.0001. Next, we continued training for 20 epochs on the LLaVA-Instruct-150K dataset with cosine learning rate annealing—both phases using image-only data. Finally, we fine-tuned on the Video-MME dataset, uniformly sampling 50 batches of frames from each video while maintaining a batch size of 8 $\times$ 8.
Once trained, we applied $G$ to generate perturbations for videos in MMBench-Video benchmark. For videos with frame counts $\leq$ 300, we processed the entire sequence in a single forward pass. For longer videos, we used sequential processing with a batch size of 300 frames, enabling perturbation generation for multi-minute videos within seconds. For the auxiliary model $\mathcal{F}$, we utilized an adversarially trained ResNet-50 on ImageNet \cite{russakovsky2015imagenet}. In equation \eqref{eq:total}, we set $\lambda_1=0.1$, $\lambda_2=20$, and $\lambda_3=10$ to balance the different types of losses.

\begin{table*}[]
\centering
\small 
\resizebox{\textwidth}{!}{
\tablestyle{6pt}{1.0}
\begin{tabular}{c|c|a|cccc|c|ccccc|c}
\shline
 \multirow{2}{*}{\bf Model} & \multirow{2}{*}{\bf \makecell{Attack}} & \cellcolor{gray!10}\bf \makecell{Overall} & \multicolumn{5}{c|}{{\cellcolor{maroon}}{\bf Perception}} &  \multicolumn{6}{c}{{\cellcolor{lightblue}}{\bf Reasoning}} \\  \cline{4-14} 
 & & {\bf Score} & \textbf{CP} & \textbf {FP-S} & \textbf {FP-C} & \textbf {HL} & \textbf {Mean} & \textbf {LR} & \textbf {AR} & \textbf {RR} & \textbf {CSR} & \textbf {TR} & \textbf {Mean} \\ \shline
\rowcolor{LIGHT_GREEN}
\multicolumn{14}{c}{\emph{Open-Source V-MLLMs}}  \\ \shline
\multirow{6}{*}{\begin{sideways}\shortstack{InternVL2.5\\8B}\end{sideways}}
& Clean & 1.61 & 1.72 & 1.76 & 1.49 & 0.56 & 1.67 & 0.96 & 1.60 & 1.77 & 1.48 & 1.38 & 1.46 \\
& GCMA & 1.59 & 1.69 & 1.73 & 1.42 & 0.65 & 1.64 & 0.95 & 1.57 & 1.77 & 1.49 & 1.34 & 1.44 \\
& CWA & 1.53 & 1.65 & 1.67 & 1.37 & 0.55 & 1.58 & 0.98 & 1.56 & 1.59 & 1.46 & 1.31 & 1.40 \\
& AnyAttack & 1.53 & 1.64 & 1.67 & 1.38 & 0.53 & 1.58 & 1.00 & 1.56 & 1.63 & 1.47 & 1.30 & 1.40 \\
& X-Transfer & 1.47 & 1.56 & 1.61 & 1.29 & 0.39 & 1.51 & 1.04 & 1.45 & 1.61 & 1.32 & 1.24 & 1.34 \\
& \textbf{CAVALRY-V} & \textbf{1.42} & 1.59 & 1.54 & 1.20 & 0.44 & 1.46 & 1.04 & 1.36 & 1.45 & 1.38 & 1.30 & 1.32 \\
\shline
\multirow{6}{*}{\begin{sideways}\shortstack{MiniCPM-o\\2.6}\end{sideways}}
& Clean & 1.72 & 1.82 & 1.87 & 1.57 & 0.84 & 1.79 & 1.46 & 1.64 & 1.64 & 1.78 & 1.55 & 1.60 \\
& GCMA & 1.70 & 1.75 & 1.84 & 1.52 & 0.76 & 1.74 & 1.49 & 1.57 & 1.73 & 1.90 & 1.52 & 1.61 \\
& CWA & 1.63 & 1.66 & 1.77 & 1.45 & 0.73 & 1.67 & 1.40 & 1.51 & 1.61 & 1.85 & 1.52 & 1.56 \\
& AnyAttack & 1.63 & 1.66 & 1.76 & 1.43 & 0.74 & 1.66 & 1.42 & 1.51 & 1.64 & 1.85 & 1.51 & 1.56 \\
& X-Transfer & 1.63 & 1.69 & 1.78 & 1.46 & 0.65 & 1.67 & 1.41 & 1.49 & 1.55 & 1.95 & 1.50 & 1.54 \\
& \textbf{CAVALRY-V} & \textbf{1.59} & 1.69 & 1.72 & 1.38 & 0.73 & 1.63 & 1.42 & 1.45 & 1.61 & 1.81 & 1.55 & 1.55 \\
\shline
\multirow{6}{*}{\begin{sideways}\shortstack{QwenVL-2.5\\7B}\end{sideways}}
& Clean & 1.45 & 1.67 & 1.45 & 1.12 & 1.13 & 1.44 & 1.60 & 1.60 & 1.55 & 1.48 & 1.17 & 1.45 \\
& GCMA & 1.43 & 1.63 & 1.41 & 1.16 & 1.13 & 1.41 & 1.66 & 1.63 & 1.43 & 1.59 & 1.15 & 1.44 \\
& CWA & 1.36 & 1.50 & 1.34 & 1.04 & 1.15 & 1.33 & 1.56 & 1.52 & 1.45 & 1.53 & 1.06 & 1.37 \\
& AnyAttack & 1.31 & 1.47 & 1.30 & 0.99 & 1.08 & 1.29 & 1.58 & 1.45 & 1.33 & 1.60 & 1.07 & 1.35 \\
& X-Transfer & 1.30 & 1.45 & 1.27 & 0.97 & 1.11 & 1.27 & 1.73 & 1.40 & 1.27 & 1.49 & 1.07 & 1.34 \\
& \textbf{CAVALRY-V} & \textbf{1.25} & 1.39 & 1.24 & 0.91 & 1.18 & 1.23 & 1.62 & 1.35 & 1.17 & 1.49 & 1.03 & 1.28 \\
\shline
\multirow{6}{*}{\begin{sideways}Aria\end{sideways}}
& Clean & 1.60 & 1.71 & 1.62 & 1.31 & 0.87 & 1.58 & 1.61 & 1.80 & 1.67 & 1.78 & 1.48 & 1.64 \\
& GCMA & 1.59 & 1.75 & 1.61 & 1.38 & 0.94 & 1.57 & 1.61 & 1.78 & 1.71 & 1.75 & 1.41 & 1.62 \\
& CWA & 1.56 & 1.67 & 1.57 & 1.37 & 1.00 & 1.53 & 1.63 & 1.80 & 1.64 & 1.75 & 1.38 & 1.61 \\
& AnyAttack & 1.52 & 1.60 & 1.55 & 1.28 & 0.66 & 1.50 & 1.59 & 1.71 & 1.55 & 1.72 & 1.38 & 1.56 \\
& X-Transfer & 1.51 & 1.58 & 1.53 & 1.28 & 0.65 & 1.48 & 1.60 & 1.65 & 1.56 & 1.83 & 1.38 & 1.56 \\
& \textbf{CAVALRY-V} & \textbf{1.50} & 1.59 & 1.53 & 1.21 & 0.74 & 1.47 & 1.60 & 1.62 & 1.58 & 1.73 & 1.36 & 1.54 \\
\shline
\multirow{6}{*}{\begin{sideways}\shortstack{Llava-Video\\7B}\end{sideways}}
& Clean & 1.58 & 1.72 & 1.71 & 1.43 & 0.23 & 1.62 & 1.27 & 1.67 & 1.42 & 1.72 & 1.35 & 1.48 \\
& GCMA & 1.56 & 1.66 & 1.71 & 1.41 & 0.16 & 1.60 & 1.28 & 1.60 & 1.42 & 1.59 & 1.33 & 1.44 \\
& CWA & 1.52 & 1.61 & 1.65 & 1.38 & 0.18 & 1.55 & 1.31 & 1.62 & 1.40 & 1.64 & 1.29 & 1.44 \\
& AnyAttack & 1.49 & 1.57 & 1.63 & 1.39 & 0.18 & 1.53 & 1.30 & 1.58 & 1.32 & 1.56 & 1.25 & 1.39 \\
& X-Transfer & 1.47 & 1.52 & 1.62 & 1.39 & 0.19 & 1.51 & 1.30 & 1.50 & 1.33 & 1.62 & 1.21 & 1.37 \\
& \textbf{CAVALRY-V} & \textbf{1.45} & 1.54 & 1.58 & 1.34 & 0.18 & 1.49 & 1.27 & 1.53 & 1.33 & 1.46 & 1.21 & 1.36 \\
\shline
\rowcolor{LIGHT_BLUE}
\multicolumn{14}{c}{\emph{Commercial V-MLLMs}}  \\ \shline
\multirow{6}{*}{\begin{sideways}\shortstack{Gemini-2.0\\Flash}\end{sideways}}
& Clean & 1.59 & 1.82 & 1.81 & 1.15 & 1.25 & 1.66 & 1.30 & 1.80 & 1.16 & 1.25 & 1.33 & 1.44 \\
& GCMA & 1.48 & 1.85 & 1.59 & 1.12 & 0.75 & 1.51 & 1.10 & 1.74 & 1.26 & 1.33 & 1.27 & 1.42 \\
& CWA & 1.48 & 1.76 & 1.62 & 1.21 & 0.75 & 1.53 & 1.50 & 1.69 & 1.05 & 1.42 & 1.20 & 1.38 \\
& AnyAttack & 1.41 & 1.67 & 1.56 & 0.88 & 1.50 & 1.44 & 1.70 & 1.51 & 1.00 & 1.83 & 1.23 & 1.40 \\
& X-Transfer & 1.44 & 1.67 & 1.68 & 1.03 & 1.00 & 1.54 & 1.10 & 1.46 & 1.11 & 1.50 & 1.37 & 1.34 \\
& \textbf{CAVALRY-V} & \textbf{1.39} & 1.55 & 1.52 & 0.97 & 0.50 & 1.40 & 1.70 & 1.43 & 1.11 & 1.58 & 1.13 & 1.35 \\
\shline
\multirow{6}{*}{\begin{sideways}\shortstack{GPT-4.1\\Mini}\end{sideways}}
& Clean & 1.88 & 2.39 & 2.01 & 1.45 & 0.25 & 1.93 & 1.70 & 1.89 & 2.00 & 2.25 & 1.63 & 1.82 \\
& GCMA & 1.80 & 2.24 & 1.85 & 1.36 & 0.25 & 1.81 & 1.70 & 1.91 & 2.05 & 2.17 & 1.47 & 1.82 \\
& CWA & 1.77 & 2.09 & 1.81 & 1.58 & 0.25 & 1.79 & 1.60 & 1.83 & 1.89 & 2.08 & 1.43 & 1.73 \\
& AnyAttack & 1.74 & 1.97 & 1.77 & 1.52 & 0.50 & 1.76 & 1.80 & 1.80 & 1.89 & 1.92 & 1.53 & 1.73 \\
& X-Transfer & 1.71 & 2.03 & 1.72 & 1.58 & 0.25 & 1.71 & 1.80 & 1.83 & 1.84 & 1.83 & 1.57 & 1.74 \\
& \textbf{CAVALRY-V} & \textbf{1.70} & 1.88 & 1.74 & 1.67 & 0.25 & 1.73 & 1.40 & 1.66 & 1.89 & 2.08 & 1.33 & 1.60 \\
\shline
\end{tabular}
}
\caption{Performance comparison on MMBench-Video benchmark. Our method (CAVALRY-V) consistently outperforms all baselines across all black-box models. Lower scores indicates more successful attacks. Best results are in \textbf{bold}.}
\label{tab:video_results}
\vspace{-10pt}
\end{table*}

\subsection{Main Results}

We evaluate CAVALRY-V against baseline attacks across both open-source and commercial V-MLLMs using the MMBench-video benchmark. Table~\ref{tab:video_results} presents comparative results for seven V-MLLMs, where lower scores indicate more successful attacks. Supplementary fine-grained results are provided in Appendix Tables~\ref{tab:main-results-fine-1} and~\ref{tab:main-results-fine-2}.
CAVALRY-V consistently outperforms all baseline methods across every tested model. Measuring performance degradation (clean score minus attack score), our method achieves \textbf{an average improvement of 22.8\% over the next best baseline} across all models. The improvements are particularly pronounced for InternVL2.5-8B (35.7\% improvement over the best baseline), QwenVL-2.5-7B (33.3\%), and MiniCPM-o-2.6 (44.4\%). Even on more robust commercial models like Gemini-2.0-Flash (11.1\%) and GPT-4.1-Mini (5.9\%), our method still provides measurable gains. The performance degradation is most significant in perception-related categories, confirming our approach effectively targets the visual-linguistic pathway. The consistent superiority across diverse architectures demonstrates the strong transferability of our attack method.

\begin{figure}[!htbp]
    \centering
    \begin{minipage}{\textwidth}
        \centering
        \includegraphics[width=\textwidth]{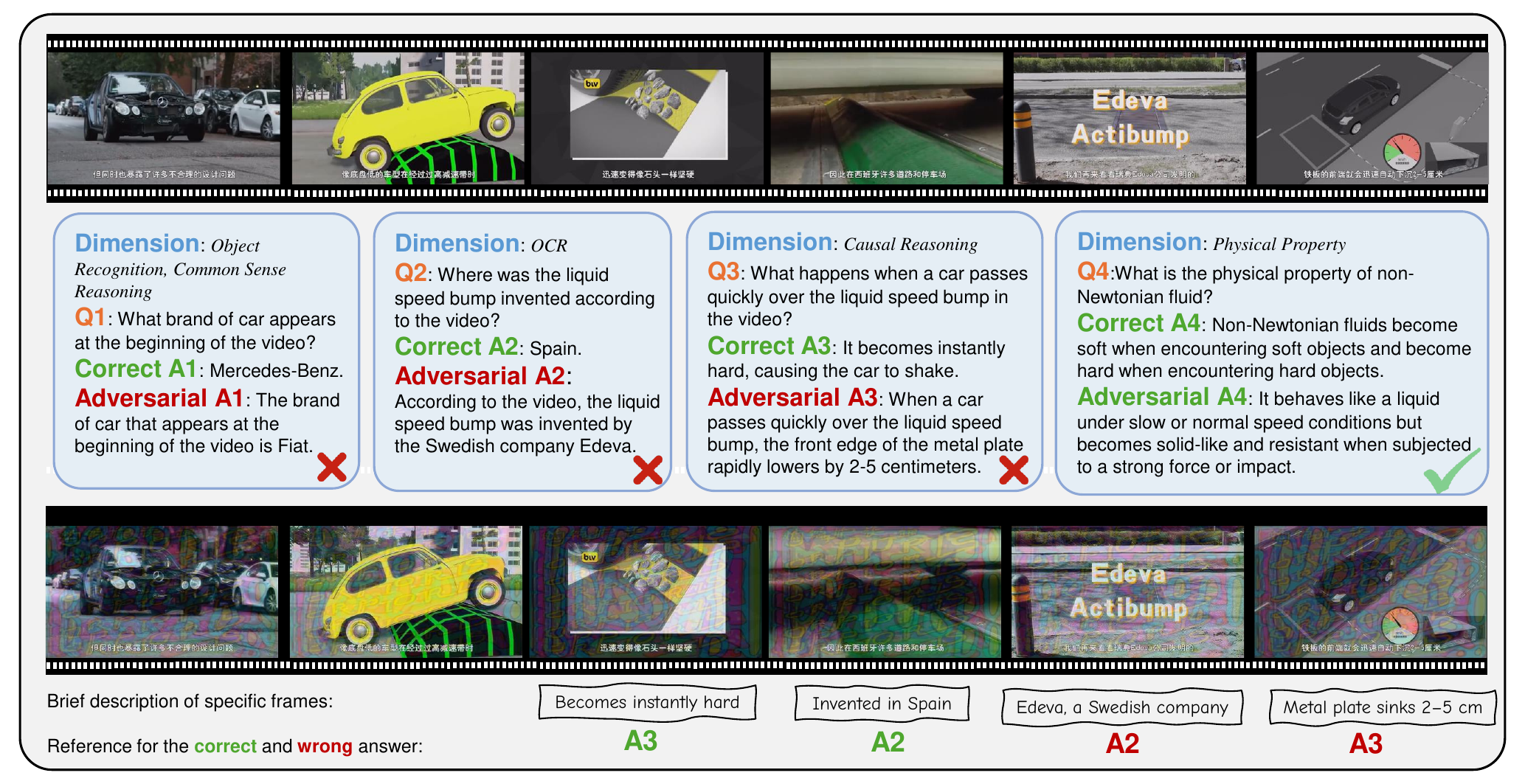}
        \caption{Visualization of attack effects on video understanding. Original video frames (top) and their adversarially perturbed versions (bottom). The attack succeeds for A1, A2 and A3, but fails for A4. Green/red text indicates correct/incorrect answers.}
        \label{fig:attack_mechanism}
    \end{minipage}
    
    
    \begin{minipage}{0.45\textwidth}
        \centering
        \includegraphics[width=\linewidth]{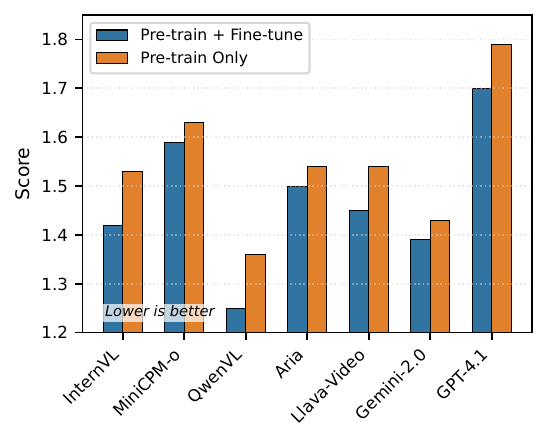}
        \caption{Performance comparison between our full pipeline and using pre-training only.}
        \label{fig:ablation}
    \end{minipage}
    \hfill
    \begin{minipage}{0.5\textwidth}
    \centering
    \begin{tabular}{lc}
        \toprule
        \textbf{Method} & \textbf{NFC} \\
        \midrule
        CWA & N/A \\
        X-Transfer & N/A \\
        AnyAttack & -4.07 \\
        GCMA & -5.99 \\
        Uniform Noise & -0.53 \\
        \textbf{CAVALRY-V} & \textbf{0.84} \\
        \bottomrule
    \end{tabular}
    \captionof{table}{Temporal consistency comparison using NFC. Higher values indicate better temporal coherence. X-Transfer and CWA are not applicable as they apply fixed perturbations across frames.}
    \label{tab:nfc_results}
    \end{minipage}
    \vspace{-15pt}
\end{figure}



\subsection{Further Analysis}

\paragraph{Attack Mechanism Analysis} To better understand how CAVALRY-V works in practice, we present a qualitative example in Figure~\ref{fig:attack_mechanism}. This example demonstrates three distinct scenarios: (1) \textit{Incorrect object recognition}, shown in A1 where a Mercedes-Benz is misidentified as a Fiat; (2) \textit{Incorrect reasoning despite accurate visual perception}, evident in A2 and A3 where the model correctly perceives visual elements but provides contextually incorrect answers; and (3) \textit{Attack failure when answers rely on stored knowledge}, demonstrated in A4 where the model correctly describes non-Newtonian fluids despite the adversarial perturbation. These observations suggest that CAVALRY-V primarily disrupts the connection between visual input and language generation. When answers depend heavily on visual evidence (A1-A3), the attack successfully breaks this connection. However, when questions can be answered using knowledge already stored in the model (A4), the attack is less effective. This analysis helps us understand both when V-MLLMs are vulnerable to attack and when they show natural resistance.

\paragraph{Ablation Study}
To evaluate our two-stage training strategy, we compared it against a one-stage variant using only LAION-400M pre-training without subsequent fine-tuning. Figure \ref{fig:ablation} shows the two-stage approach consistently outperforms across all models. This confirms that while large-scale pre-training establishes foundational adversarial patterns, the fine-tuning phase is essential for adapting these patterns to spatiotemporal coherence. The consistent improvement demonstrates the effectiveness of our two-stage training strategy.

\paragraph{Temporal Consistency Analysis} 
We evaluate the temporal consistency of perturbations using the Normalized Flow Consistency (NFC) metric, which measures how well perturbations adhere to the video's motion patterns. Mathematically, NFC is defined as $\text{NFC} = 1 - \frac{\|W(\delta_{t-1}) - \delta_t\|_2}{\|\delta_t\|_2}$, where $W(\cdot)$ is the warping operation based on the optical flow between consecutive frames. NFC ranges from (-$\infty$, 1], with higher values indicating better consistency. Table~\ref{tab:nfc_results} presents the comparison results. Our method achieves a positive NFC score, demonstrating its ability to generate perturbations that coherently follow the temporal dynamics of the video. In contrast, all baseline methods produce negative NFC values, indicating their perturbations move contradictory to the video content. These results validate that our approach successfully produces consistent perturbations across temporal sequences.

\begin{table*}[]
\centering
\small 
\resizebox{\textwidth}{!}{
\tablestyle{3pt}{1.2}
\begin{tabular}{c|c|a|cccccccccc|cccc}
\shline
 \multirow{2}{*}{\bf Model} & \multirow{2}{*}{\bf \makecell{Attack}} & {\bf \makecell{Overall}} & \multicolumn{10}{c|}{{\cellcolor{maroon}}{\bf Perception}} &  \multicolumn{4}{c}{{\cellcolor{lightblue}}{\bf Reasoning}} \\  \cline{4-17} 
 & & {\bf Score} & \textbf{OCR} & \textbf{Art.} & \textbf{Celeb.} & \textbf{Color} & \textbf{Count} & \textbf{Exist.} & \textbf{Land.} & \textbf{Posi.} & \textbf{Posters} & \textbf{Scene} & \textbf{Code.} & \textbf{Comm.} & \textbf{Num.} & \textbf{Text.} \\ \shline
\rowcolor{LIGHT_GREEN}
\multicolumn{17}{c}{\emph{Open-Source V-MLLMs}}  \\ \shline

\multirow{5}{*}{\begin{sideways}\shortstack{InternVL2.5\\8B}\end{sideways}} 
& Clean & 2310.0 & 180.0 & 148.5 & 111.8 & 160.0 & 173.3 & 195.0 & 171.0 & 163.3 & 166.3 & 163.8 & 160.0 & 144.3 & 145.0 & 202.5 \\
& CWA & 2210.5 & 175.0 & 141.3 & 107.1 & 154.2 & 166.7 & 185.0 & 165.0 & 156.7 & 161.6 & 158.8 & 150.0 & 139.3 & 135.0 & 195.0 \\
& AnyAttack & 2235.6 & 177.5 & 144.8 & 108.2 & 156.7 & 170.0 & 190.0 & 167.5 & 160.0 & 163.3 & 161.3 & 155.0 & 141.4 & 140.0 & 200.0 \\
& X-Transfer & 2198.4 & 177.5 & 142.8 & 109.4 & 158.3 & 168.3 & 190.0 & 166.0 & 158.3 & 162.6 & 159.3 & 157.5 & 142.9 & 142.5 & 197.5 \\
& \textbf{CAVALRY} & \textbf{2163.6} & 170.0 & 138.0 & 105.9 & 151.7 & 163.3 & 180.0 & 162.5 & 153.3 & 159.9 & 156.3 & 147.5 & 137.1 & 132.5 & 192.5 \\ \hline

\multirow{5}{*}{\begin{sideways}\shortstack{MiniCPM-o\\2.6}\end{sideways}} 
& Clean & 2266.8 & 147.5 & 146.5 & 151.5 & 188.3 & 173.3 & 200.0 & 172.8 & 143.3 & 178.6 & 160.8 & 130.0 & 149.3 & 140.0 & 185.0 \\
& CWA & 2124.7 & 143.8 & 119.8 & 111.5 & 155.0 & 150.0 & 185.0 & 170.0 & 140.0 & 177.6 & 150.3 & 137.5 & 137.1 & 162.5 & 180.0 \\
& AnyAttack & 2173.7 & 147.5 & 122.0 & 113.5 & 158.3 & 153.3 & 190.0 & 173.0 & 143.3 & 179.9 & 152.8 & 145.0 & 140.0 & 170.0 & 185.0 \\
& X-Transfer & 2118.8 & 142.5 & 121.8 & 112.1 & 156.7 & 151.7 & 190.0 & 171.5 & 141.7 & 178.9 & 151.3 & 145.0 & 141.4 & 167.5 & 182.5 \\
& \textbf{CAVALRY} & \textbf{2052.1} & 140.0 & 117.3 & 109.4 & 151.7 & 146.7 & 180.0 & 167.3 & 136.7 & 176.2 & 147.8 & 132.5 & 134.3 & 155.0 & 177.5 \\ \hline

\multirow{5}{*}{\begin{sideways}\shortstack{QwenVL2.5\\7B}\end{sideways}} 
& Clean & 2273.0 & 187.5 & 125.5 & 104.4 & 183.3 & 161.7 & 190.0 & 170.0 & 148.3 & 173.5 & 159.3 & 167.5 & 130.0 & 152.5 & 187.5 \\
& CWA & 2190.8 & 182.5 & 120.3 & 99.4 & 176.7 & 155.0 & 180.0 & 164.0 & 141.7 & 167.7 & 153.8 & 157.5 & 125.0 & 142.5 & 180.0 \\
& AnyAttack & 2191.6 & 185.0 & 122.8 & 100.9 & 180.0 & 158.3 & 185.0 & 166.5 & 145.0 & 169.7 & 156.3 & 162.5 & 127.1 & 147.5 & 185.0 \\
& X-Transfer & 2167.1 & 182.5 & 121.3 & 102.4 & 178.3 & 156.7 & 185.0 & 165.3 & 143.3 & 168.7 & 154.3 & 162.5 & 127.9 & 145.0 & 182.5 \\
& \textbf{CAVALRY} & \textbf{2143.1} & 180.0 & 117.3 & 98.8 & 173.3 & 151.7 & 175.0 & 161.8 & 138.3 & 165.7 & 151.3 & 155.0 & 122.9 & 140.0 & 182.5 \\ \hline

\multirow{5}{*}{\begin{sideways}Aria\end{sideways}} 
& Clean & 2217.1 & 170.0 & 143.5 & 140.6 & 185.0 & 168.3 & 190.0 & 169.5 & 156.7 & 179.6 & 156.8 & 117.5 & 137.1 & 117.5 & 185.0 \\
& CWA & 2116.0 & 160.0 & 137.5 & 127.9 & 165.0 & 163.3 & 175.0 & 169.5 & 141.7 & 177.9 & 163.5 & 107.5 & 137.1 & 112.5 & 177.5 \\
& AnyAttack & 2108.0 & 162.5 & 132.5 & 110.6 & 153.3 & 156.7 & 180.0 & 163.3 & 158.3 & 179.3 & 158.8 & 122.5 & 132.9 & 120.0 & 177.5 \\
& X-Transfer & 2129.6 & 162.5 & 133.8 & 108.8 & 153.3 & 160.0 & 185.0 & 161.3 & 158.3 & 176.2 & 155.8 & 137.5 & 137.1 & 115.0 & 185.0 \\
& \textbf{CAVALRY} & \textbf{2037.3} & 155.0 & 128.0 & 102.9 & 145.0 & 163.3 & 175.0 & 156.8 & 146.7 & 178.2 & 161.8 & 100.0 & 132.1 & 107.5 & 185.0 \\ \hline

\multirow{5}{*}{\begin{sideways}\shortstack{LLaVA-Video\\7B}\end{sideways}} 
& Clean & 1937.5 & 85.0 & 127.5 & 119.1 & 156.7 & 168.3 & 195.0 & 152.0 & 146.7 & 170.4 & 170.0 & 110.0 & 137.1 & 105.0 & 90.0 \\
& CWA & 1853.2 & 77.5 & 122.0 & 114.1 & 150.0 & 161.7 & 185.0 & 145.5 & 140.0 & 165.0 & 164.0 & 102.5 & 131.4 & 97.5 & 82.5 \\
& AnyAttack & 1878.7 & 80.0 & 123.8 & 116.2 & 153.3 & 165.0 & 190.0 & 148.0 & 143.3 & 167.4 & 167.5 & 105.0 & 134.3 & 100.0 & 85.0 \\
& X-Transfer & 1861.4 & 82.5 & 125.3 & 117.1 & 151.7 & 163.3 & 190.0 & 146.3 & 141.7 & 165.3 & 166.0 & 107.5 & 135.7 & 102.5 & 87.5 \\
& \textbf{CAVALRY} & \textbf{1806.4} & 75.0 & 120.0 & 112.4 & 146.7 & 158.3 & 180.0 & 142.8 & 136.7 & 162.9 & 160.5 & 100.0 & 128.6 & 95.0 & 80.0 \\ \hline

\rowcolor{LIGHT_BLUE}
\multicolumn{17}{c}{\emph{Commercial V-MLLMs}}  \\ \shline

\multirow{5}{*}{\begin{sideways}\shortstack{Gemini-2.0\\Flash}\end{sideways}} 
& Clean & 2394.7 & 185.0 & 150.0 & 160.0 & 190.0 & 173.3 & 200.0 & 169.0 & 113.3 & 181.0 & 153.8 & 185.0 & 164.3 & 177.5 & 192.5 \\
& CWA & 2310.6 & 200.0 & 150.3 & 122.9 & 180.0 & 150.0 & 185.0 & 168.3 & 118.3 & 185.0 & 160.8 & 185.0 & 150.0 & 162.5 & 192.5 \\
& AnyAttack & 2252.1 & 185.0 & 141.0 & 105.0 & 165.0 & 148.3 & 195.0 & 169.3 & 123.3 & 183.7 & 157.3 & 177.5 & 139.3 & 192.5 & 170.0 \\
& X-Transfer & \textbf{2225.8} & 185.0 & 132.5 & 105.0 & 165.0 & 140.0 & 195.0 & 159.8 & 125.0 & 177.2 & 154.5 & 185.0 & 139.3 & 185.0 & 177.5 \\
& \textbf{CAVALRY} & 2243.5 & 192.5 & 138.8 & 102.4 & 145.0 & 153.3 & 195.0 & 161.5 & 135.0 & 182.0 & 161.3 & 185.0 & 144.3 & 177.5 & 170.0 \\ \hline

\multirow{5}{*}{\begin{sideways}\shortstack{GPT4.1\\Mini}\end{sideways}} 
& Clean & 2122.4 & 185.0 & 121.5 & 71.2 & 183.3 & 170.0 & 190.0 & 123.5 & 148.3 & 164.0 & 140.3 & 162.5 & 167.9 & 155.0 & 140.0 \\
& CWA & 2001.1 & 185.0 & 108.8 & 57.9 & 193.3 & 153.3 & 175.0 & 120.3 & 141.7 & 161.9 & 146.8 & 132.5 & 152.1 & 140.0 & 132.5 \\
& AnyAttack & 1953.8 & 192.5 & 103.8 & 56.2 & 183.3 & 143.3 & 180.0 & 102.0 & 126.7 & 162.3 & 141.3 & 140.0 & 150.0 & 140.0 & 132.5 \\
& X-Transfer & 1961.0 & 192.5 & 98.5 & 56.2 & 170.0 & 165.0 & 185.0 & 97.8 & 138.3 & 163.6 & 133.8 & 125.0 & 147.9 & 155.0 & 132.5 \\
& \textbf{CAVALRY} & \textbf{1899.8} & 185.0 & 100.3 & 52.7 & 145.0 & 153.3 & 195.0 & 95.0 & 121.7 & 165.3 & 137.0 & 140.0 & 137.1 & 140.0 & 132.5 \\ 
\shline
\end{tabular}
}
\caption{Performance comparison on the MME image benchmark. Our method (CAVALRY) consistently achieves the strongest attack performance despite being primarily designed for video data. Lower scores indicates more successful attacks. Best results are in \textbf{bold}.}
\label{tab:image_results}
\vspace{-15pt}
\end{table*}

\subsection{Generalization to Image Attacks}


While CAVALRY-V is designed for videos, the underlying approach can be naturally adapted to single-frame scenarios, denoted as CAVALRY. We evaluated its generalization to the MME image benchmark (2,374 questions spanning perception and reasoning dimensions). Table~\ref{tab:image_results} shows our approach outperforms baselines on six out of seven models, with Gemini-2.0-Flash as the only exception. Quantitatively, measuring performance degradation (clean score minus attack score), CAVALRY achieves \textbf{an average improvement of 34.4\% over the best baseline attacks} across all models. The improvements range from 22.7\% (QwenVL2.5-7B) to 64.8\% (Aria), while for Gemini-2.0-Flash, our method underperforms the best baseline by 10.5\%. This strong cross-domain performance demonstrates our core mechanism for disrupting cross-modal integration transfers effectively to static images, despite being primarily optimized for video data.

\section{Conclusion}

We presented CAVALRY-V, a transferable generator framework for adversarial attacks on V-MLLMs. Our approach introduced two key innovations: (1) a cross-modal disruption mechanism targeting the interface between visual perception and language generation, and (2) a progressive multi-stage training strategy combining large-scale foundation pretraining with visual-linguistic and temporal coherence fine-tuning. Extensive experiments demonstrated that our method consistently outperforms existing approaches across diverse V-MLLM architectures. It represents a significant step toward understanding and addressing the vulnerabilities of emerging multimodal systems.


\bibliography{main}
\bibliographystyle{plainnat}


\newpage

\appendix

\section{Limitations}
The computational demands of our approach, particularly during the foundation pretraining stage on LAION-400M, create significant barriers to implementation in resource-constrained environments. To mitigate this issue, we have publicly released our model weights, enabling researchers to build upon our work without needing to replicate the intensive pretraining process.
These high computational requirements also make comprehensive ablation studies challenging to conduct. As a result, hyperparameters were primarily designed manually based on empirical observations rather than through exhaustive parameter sweeps or optimization. This limitation potentially leaves room for further performance improvements through more systematic parameter tuning.
The dual memory demands of both V-MLLMs and adversarial attack generation posed additional constraints. When deploying our surrogate model, we were limited to using the 1B parameter variant of InternVL-2.5 due to GPU memory constraints. We anticipate that utilizing larger surrogate models would likely yield even stronger attack performance, as more powerful models could potentially generate more effective and transferable adversarial patterns.

\section{Broader Impacts}
Our work on adversarial attacks for Video MLLMs has several societal implications. On the positive side, CAVALRY-V enables preemptive identification of robustness vulnerabilities in these models before widespread deployment, allowing developers to implement necessary safeguards. This work advances understanding of cross-modal security challenges and establishes evaluation benchmarks for robust multimodal systems.
However, we acknowledge potential negative impacts. Our attack techniques could be misused to generate deceptive content or undermine V-MLLM reliability.
To mitigate these concerns, we have followed responsible disclosure practices. When releasing model weights, we will implement access controls requiring researchers to agree to usage guidelines that discourage harmful applications. We emphasize that understanding these vulnerabilities is crucial for developing more robust systems as V-MLLMs continue deployment in increasingly critical domains where reliability is paramount.

\begin{table*}[]
\centering
\small 
\tablestyle{3pt}{1.1}
\begin{tabular}{ll|cccccccccccccc}
\shline
\textbf{Model} & \textbf{Method} & \textbf{VT} & \textbf{VE} & \textbf{VS} & \textbf{VSt} & \textbf{OCR} & \textbf{OR} & \textbf{AR} & \textbf{ER} & \textbf{HM} & \textbf{C} & \textbf{SR} & \textbf{HOI} & \textbf{HI} \\ \shline
\multirow{6}{*}{\begin{sideways}\shortstack{InternVL2.5\\8B}\end{sideways}}  
 & Clean & 1.71 & 1.65 & 1.70 & 1.93 & 1.79 & 1.81 & 2.04 & 1.49 & 1.44 & 1.78 & 1.67 & 1.41 & 1.49 \\
 & GCMA & 1.68 & 1.65 & 1.63 & 1.90 & 1.80 & 1.71 & 1.96 & 1.37 & 1.28 & 1.82 & 1.60 & 1.35 & 1.41 \\
 & CWA & 1.69 & 1.53 & 1.71 & 1.60 & 1.77 & 1.64 & 1.90 & 1.34 & 1.24 & 1.65 & 1.42 & 1.32 & 1.46 \\
 & Any-attack & 1.66 & 1.53 & 1.68 & 1.64 & 1.77 & 1.65 & 1.90 & 1.34 & 1.20 & 1.64 & 1.47 & 1.32 & 1.41 \\
 & X-Transfer & 1.61 & 1.51 & 1.54 & 1.52 & 1.71 & 1.51 & 1.87 & 1.36 & 1.11 & 1.64 & 1.53 & 1.19 & 1.28 \\ 
 & \textbf{CAVALRY-V} & 1.64 & 1.51 & 1.56 & 1.69 & 1.70 & 1.37 & 1.76 & 1.35 & 1.10 & 1.64 & 1.38 & 1.10 & 1.28 \\
 \shline

\multirow{6}{*}{\begin{sideways}\shortstack{MiniCPM-o\\2.6}\end{sideways}} 
 & Clean & 1.82 & 1.73 & 1.93 & 1.83 & 2.04 & 1.92 & 2.06 & 1.52 & 1.39 & 1.66 & 1.71 & 1.52 & 1.57 \\
 & GCMA & 1.81 & 1.62 & 1.77 & 1.86 & 1.99 & 1.82 & 2.05 & 1.58 & 1.39 & 1.59 & 1.60 & 1.49 & 1.54 \\
 & CWA & 1.71 & 1.65 & 1.67 & 1.50 & 1.99 & 1.69 & 1.87 & 1.40 & 1.29 & 1.54 & 1.60 & 1.37 & 1.48\\
 & Any-attack & 1.72 & 1.63 & 1.70 & 1.45 & 1.99 & 1.69 & 1.87 & 1.40 & 1.27 & 1.54 & 1.64 & 1.32 & 1.48\\
 & X-Transfer & 1.81 & 1.58 & 1.79 & 1.38 & 1.99 & 1.70 & 1.91 & 1.43 & 1.31 & 1.53 & 1.42 & 1.49 & 1.44 \\ 
 & \textbf{CAVALRY-V} & 1.81 & 1.56 & 1.74 & 1.43 & 1.95 & 1.56 & 1.89 & 1.39 & 1.34 & 1.49 & 1.51 & 1.28 & 1.51 \\
 \shline

\multirow{6}{*}{\begin{sideways}\shortstack{Qwen2.5-VL\\7B}\end{sideways}} 
 & Clean & 1.92 & 1.38 & 1.40 & 2.10 & 1.56 & 1.35 & 1.77 & 1.20 & 1.08 & 1.29 & 1.36 & 1.06 & 1.02 \\
 & GCMA & 1.90 & 1.42 & 1.26 & 2.14 & 1.54 & 1.25 & 1.77 & 1.15 & 1.08 & 1.30 & 1.24 & 1.15 & 1.10 \\
 & CWA & 1.83 & 1.34 & 1.02 & 2.00 & 1.51 & 1.16 & 1.68 & 1.12 & 0.98 & 1.22 & 1.36 & 0.89 & 1.05 \\
 & Any-attack & 1.78 & 1.27 & 1.11 & 1.88 & 1.50 & 1.07 & 1.58 & 0.99 & 1.02 & 1.14 & 0.98 & 0.95 & 1.05 \\
 & X-Transfer & 1.83 & 1.34 & 0.93 & 1.76 & 1.45 & 1.08 & 1.55 & 1.05 & 1.00 & 1.12 & 1.11 & 0.88 & 1.00 \\ 
 & \textbf{CAVALRY-V} & 1.79 & 1.19 & 0.94 & 1.76 & 1.46 & 1.02 & 1.39 & 1.07 & 0.82 & 1.12 & 1.22 & 0.77 & 0.89 \\
 \shline

\multirow{6}{*}{\begin{sideways}\shortstack{Aria}\end{sideways}} 
 & Clean & 1.80 & 1.63 & 1.54 & 2.00 & 1.60 & 1.67 & 1.95 & 1.43 & 1.48 & 1.47 & 1.49 & 1.29 & 1.26 \\
 & GCMA & 1.78 & 1.75 & 1.59 & 1.95 & 1.54 & 1.66 & 1.90 & 1.39 & 1.34 & 1.62 & 1.53 & 1.41 & 1.23 \\
 & CWA & 1.69 & 1.72 & 1.44 & 1.88 & 1.53 & 1.56 & 1.88 & 1.42 & 1.39 & 1.46 & 1.47 & 1.40 & 1.28 \\
 & Any-attack & 1.66 & 1.67 & 1.26 & 1.93 & 1.57 & 1.50 & 1.80 & 1.36 & 1.30 & 1.48 & 1.38 & 1.27 & 1.28 \\
 & X-Transfer & 1.66 & 1.71 & 1.28 & 1.74 & 1.54 & 1.42 & 1.89 & 1.34 & 1.38 & 1.55 & 1.13 & 1.37 & 1.20 \\ 
 & \textbf{CAVALRY-V} & 1.66 & 1.65 & 1.35 & 1.79 & 1.52 & 1.42 & 1.86 & 1.39 & 1.39 & 1.43 & 1.07 & 1.23 & 1.26 \\
 \shline

\multirow{6}{*}{\begin{sideways}\shortstack{LLaVA-Video\\7b}\end{sideways}} 
 & Clean & 1.58 & 1.77 & 1.79 & 2.00 & 1.74 & 1.76 & 1.90 & 1.60 & 1.33 & 1.64 & 1.40 & 1.50 & 1.34 \\
 & GCMA & 1.55 & 1.75 & 1.76 & 1.67 & 1.74 & 1.76 & 1.91 & 1.58 & 1.41 & 1.67 & 1.51 & 1.45 & 1.28 \\
 & CWA & 1.49 & 1.71 & 1.62 & 1.79 & 1.70 & 1.68 & 1.86 & 1.48 & 1.37 & 1.59 & 1.38 & 1.41 & 1.39\\
 & Any-attack & 1.44 & 1.67 & 1.55 & 1.76 & 1.70 & 1.63 & 1.79 & 1.40 & 1.25 & 1.67 & 1.44 & 1.43 & 1.33 \\
 & X-Transfer & 1.32 & 1.70 & 1.55 & 1.57 & 1.69 & 1.60 & 1.83 & 1.43 & 1.34 & 1.58 & 1.47 & 1.41 & 1.33 \\ 
 & \textbf{CAVALRY-V} & 1.41 & 1.61 & 1.60 & 1.69 & 1.68 & 1.49 & 1.78 & 1.39 & 1.26 & 1.62 & 1.47 & 1.31 & 1.31 \\
 \shline

 \multirow{6}{*}{\begin{sideways}\shortstack{Gemini-2.0\\Flash}\end{sideways}} 
 & Clean & 2.36 & 0.50 & 1.71 & 2.00 & 1.69 & 1.71 & 2.40 & 1.50 & 1.67 & 1.68 & 1.11 & 1.17 & 1.31 \\
 & GCMA & 2.43 & 0.67 & 1.43 & 2.14 & 1.47 & 1.39 & 2.33 & 1.29 & 1.83 & 1.45 & 1.11 & 1.00 & 1.31  \\
 & CWA & 2.14 & 0.67 & 1.57 & 2.14 & 1.44 & 1.52 & 2.40 & 1.57 & 1.33 & 1.45 & 1.44 & 1.25 & 1.15\\
 & Any-attack &  2.36 & 0.50 & 1.43 & 1.43 & 1.29 & 1.35 & 2.20 & 1.57 & 1.50 & 1.73 & 0.44 & 1.00 & 1.00\\
 & X-Transfer & 2.07 & 0.33 & 1.71 & 1.86 & 1.49 & 1.45 & 2.13 & 1.50 & 1.00 & 2.00 & 0.67 & 1.17 & 1.08 \\ 
 & \textbf{CAVALRY-V} & 2.14 & 0.33 & 1.43 & 1.57 & 1.31 & 1.29 & 2.13 & 1.29 & 1.17 & 1.77 & 0.89 & 1.17 & 0.77  \\
 \shline

   \multirow{6}{*}{\begin{sideways}\shortstack{GPT4.1\\Mini}\end{sideways}} 
 & Clean & 2.64 & 1.33 & 2.43 & 2.71 & 2.16 & 1.52 & 2.47 & 2.00 & 1.50 & 2.00 & 2.11 & 1.17 & 1.38  \\
 & GCMA & 2.50 & 1.33 & 2.00 & 2.71 & 1.87 & 1.48 & 2.40 & 1.57 & 1.67 & 1.82 & 2.22 & 1.00 & 1.23\\
 & CWA & 2.43 & 1.33 & 1.57 & 2.57 & 2.02 & 1.35 & 2.47 & 1.57 & 1.17 & 1.50 & 2.22 & 1.25 & 1.54 \\
 & Any-attack & 2.29 & 1.17 & 1.29 & 2.71 & 1.80 & 1.48 & 2.47 & 1.79 & 1.00 & 1.50 & 2.00 & 1.17 & 1.62 \\
 & X-Transfer & 2.14 & 1.17 & 1.86 & 2.71 & 1.73 & 1.39 & 2.13 & 1.71 & 1.17 & 1.64 & 2.22 & 1.17 & 1.62\\ 
 & \textbf{CAVALRY-V} & 2.14 & 0.83 & 1.57 & 2.57 & 1.73 & 1.48 & 2.27 & 1.50 & 1.17 & 1.77 & 2.11 & 1.42 & 1.69\\
 \shline
 
\end{tabular}
\caption{VT: Video Topic, VE: Video Emotion, VS: Video Scene, VSt: Video Style, OCR: OCR, OR: Object Recognition, AR: Attribute Recognition, ER: Event Recognition, HM: Human Motion, C: Counting, SR: Spatial Relationship, HOI: Human-object Interaction, HI: Human Interaction.}
\label{tab:main-results-fine-1}
\end{table*}

\begin{table*}[]
\centering
\small
\tablestyle{3pt}{1.1}
\begin{tabular}{ll|cccccccccccccc}
\shline
\textbf{Model} & \textbf{Method} & \textbf{Hal} & \textbf{SIT} & \textbf{MC} & \textbf{PP} & \textbf{FR} & \textbf{IR} & \textbf{NR} & \textbf{PR} & \textbf{SR} & \textbf{CSR} & \textbf{CFR} & \textbf{CR} & \textbf{FP} \\ \shline
\multirow{6}{*}{\begin{sideways}\shortstack{InternVL2.5\\8B}\end{sideways}}
& Clean & 0.56 & 1.12 & 0.71 & 1.48 & 1.45 & 1.87 & 1.56 & 1.65 & 2.00 & 1.48 & 1.52 & 1.41 & 1.23 \\
& GCMA & 0.65 & 1.09 & 0.73 & 1.41 & 1.38 & 1.91 & 1.48 & 1.65 & 2.02 & 1.49 & 1.43 & 1.37 & 1.28 \\
& CWA & 0.55 & 1.04 & 0.89 & 1.45 & 1.36 & 1.92 & 1.41 & 1.41 & 1.89 & 1.46 & 1.44 & 1.34 & 1.26 \\
& Any-attack & 0.53 & 1.07 & 0.89 & 1.44 & 1.36 & 1.87 & 1.44 & 1.47 & 1.87 & 1.47 & 1.38 & 1.34 & 1.21 \\
& X-Transfer & 0.39 & 1.16 & 0.87 & 1.39 & 1.33 & 1.62 & 1.44 & 1.41 & 1.89 & 1.32 & 1.43 & 1.24 & 1.15 \\
& \textbf{CAVALRY-V} & 0.44 & 1.21 & 0.80 & 1.41 & 1.16 & 1.49 & 1.26 & 1.31 & 1.67 & 1.38 & 1.35 & 1.34 & 1.23 \\
\shline
\multirow{6}{*}{\begin{sideways}\shortstack{MiniCPM-o\\2.6}\end{sideways}}
& Clean & 0.84 & 1.65 & 1.18 & 1.67 & 1.53 & 1.72 & 1.74 & 1.37 & 1.85 & 1.78 & 1.45 & 1.56 & 1.53 \\
& GCMA & 0.76 & 1.71 & 1.16 & 1.54 & 1.55 & 1.60 & 1.89 & 1.55 & 1.83 & 1.90 & 1.48 & 1.57 & 1.49 \\
& CWA & 0.73 & 1.50 & 1.27 & 1.52 & 1.42 & 1.62 & 1.78 & 1.51 & 1.69 & 1.85 & 1.43 & 1.57 & 1.51\\
& Any-attack & 0.74 & 1.50 & 1.32 & 1.54 & 1.42 & 1.60 & 1.78 & 1.53 & 1.73 & 1.85 & 1.40 & 1.56 & 1.53 \\
& X-Transfer & 0.65 & 1.50 & 1.27 & 1.44 & 1.38 & 1.64 & 1.74 & 1.43 & 1.57 & 1.95 & 1.52 & 1.52 & 1.45 \\
& \textbf{CAVALRY-V} & 0.73 & 1.59 & 1.16 & 1.48 & 1.40 & 1.45 & 1.74 & 1.49 & 1.67 & 1.81 & 1.62 & 1.62 & 1.56 \\
\shline
\multirow{6}{*}{\begin{sideways}\shortstack{Qwen2.5-VL\\7B}\end{sideways}}
& Clean & 1.13 & 1.60 & 1.60 & 1.80 & 1.36 & 1.64 & 1.85 & 1.59 & 1.35 & 1.48 & 1.25 & 1.17 & 1.13 \\
& GCMA & 1.13 & 1.65 & 1.69 & 1.61 & 1.81 & 1.81 & 1.33 & 1.33 & 1.59 & 1.59 & 1.30 & 1.12 & 1.09 \\
& CWA & 1.15 & 1.41 & 1.78 & 1.56 & 1.42 & 1.58 & 1.81 & 1.41 & 1.31 & 1.53 & 1.27 & 1.02 & 0.98 \\
& Any-attack & 1.08 & 1.46 & 1.76 & 1.48 & 1.36 & 1.49 & 1.74 & 1.37 & 1.07 & 1.60 & 1.25 & 1.01 & 0.99 \\
& X-Transfer & 1.11 & 1.59 & 1.96 & 1.35 & 1.40 & 1.43 & 1.81 & 1.24 & 1.02 & 1.49 & 1.12 & 1.06 & 1.09 \\
& \textbf{CAVALRY-V} & 1.18 & 1.44 & 1.89 & 1.46 & 1.27 & 1.32 & 1.67 & 1.27 & 0.83 & 1.49 & 1.25 & 0.99 & 1.04 \\
\shline
\multirow{6}{*}{\begin{sideways}\shortstack{Aria}\end{sideways}}
& Clean & 0.87 & 1.72 & 1.44 & 1.80 & 1.78 & 1.79 & 1.96 & 1.43 & 1.76 & 1.78 & 1.43 & 1.51 & 1.45 \\
& GCMA & 0.94 & 1.75 & 1.40 & 1.63 & 1.75 & 1.98 & 1.74 & 1.59 & 1.81 & 1.75 & 1.38 & 1.43 & 1.38 \\
& CWA & 1.00 & 1.71 & 1.51 & 1.74 & 1.71 & 1.94 & 1.70 & 1.45 & 1.80 & 1.75 & 1.45 & 1.45 & 1.47 \\
& Any-attack & 0.66 & 1.68 & 1.47 & 1.67 & 1.69 & 1.75 & 1.70 & 1.39 & 1.63 & 1.72 & 1.35 & 1.41 & 1.34 \\
& X-Transfer & 0.65 & 1.76 & 1.36 & 1.56 & 1.65 & 1.74 & 1.78 & 1.31 & 1.69 & 1.83 & 1.45 & 1.32 & 1.51 \\
& \textbf{CAVALRY-V} & 0.74 & 1.65 & 1.53 & 1.57 & 1.49 & 1.79 & 1.63 & 1.47 & 1.65 & 1.73 & 1.43 & 1.35 & 1.34 \\
\shline
\multirow{6}{*}{\begin{sideways}\shortstack{LLaVA-Video\\7b}\end{sideways}}
& Clean & 0.23 & 1.41 & 1.07 & 1.67 & 1.69 & 1.64 & 1.59 & 1.41 & 1.35 & 1.72 & 1.18 & 1.47 & 1.36 \\
& GCMA & 0.16 & 1.46 & 1.02 & 1.57 & 1.62 & 1.60 & 1.67 & 1.39 & 1.31 & 1.59 & 1.27 & 1.45 & 1.47 \\
& CWA & 0.18 & 1.37 & 1.22 & 1.59 & 1.47 & 1.83 & 1.63 & 1.39 & 1.35 & 1.64 & 1.18 & 1.30 & 1.47\\
& Any-attack & 0.18 & 1.38 & 1.18 & 1.57 & 1.56 & 1.58 & 1.52 & 1.33 & 1.20 & 1.56 & 1.18 & 1.25 & 1.34 \\
& X-Transfer & 0.19 & 1.41 & 1.13 & 1.50 & 1.49 & 1.49 & 1.78 & 1.24 & 1.19 & 1.62 & 1.20 & 1.20 & 1.19 \\
& \textbf{CAVALRY-V} & 0.18 & 1.37 & 1.13 & 1.48 & 1.55 & 1.55 & 1.56 & 1.33 & 1.20 & 1.46 & 1.18 & 1.18 & 1.36 \\
\shline

 \multirow{6}{*}{\begin{sideways}\shortstack{Gemini-2.0\\Flash}\end{sideways}} 
 & Clean & 1.25 & 1.71 & 0.33 & 1.82 & 1.71 & 1.90 & 1.00 & 1.14 & 1.22 & 1.25 & 2.20 & 1.17 & 1.14 \\
 & GCMA & 0.75 & 1.57 & 0.00 & 1.91 & 1.57 & 1.80 & 1.00 & 1.29 & 1.33 & 1.33 & 2.00 & 1.17 & 1.00\\
 & CWA & 0.75 & 1.57 & 1.33 & 1.64 & 1.79 & 1.60 & 0.67 & 1.71 & 0.67 & 1.42 & 1.60 & 1.28 & 0.71\\
 & Any-attack &  1.50 & 1.57 & 2.00 & 1.64 & 1.36 & 1.60 & 0.67 & 1.57 & 0.67 & 1.83 & 1.80 & 1.17 & 1.00\\
 & X-Transfer & 1.00 & 1.57 & 0.00 & 1.27 & 1.43 & 1.70 & 1.00 & 1.57 & 0.78 & 1.50 & 2.20 & 1.22 & 1.14 \\ 
 & \textbf{CAVALRY-V} &  0.50 & 1.57 & 2.00 & 1.45 & 1.50 & 1.30 & 1.00 & 1.57 & 0.78 & 1.58 & 1.80 & 1.11 & 0.71\\
 \shline

  \multirow{6}{*}{\begin{sideways}\shortstack{GPT4.1\\Mini}\end{sideways}} 
 & Clean & 0.25 & 2.14 & 0.67 & 2.18 & 1.71 & 1.80 & 2.00 & 2.29 & 1.78 & 2.25 & 1.80 & 1.72 & 1.29\\
 & GCMA & 0.25 & 2.00 & 1.00 & 2.00 & 1.86 & 1.90 & 2.00 & 2.29 & 1.89 & 2.17 & 1.60 & 1.56 & 1.14\\
 & CWA & 0.25 & 2.00 & 0.67 & 2.09 & 1.64 & 1.80 & 2.00 & 2.14 & 1.67 & 2.08 & 1.20 & 1.61 & 1.14\\
 & Any-attack &  0.50 & 2.14 & 1.00 & 2.00 & 1.71 & 1.70 & 2.00 & 2.14 & 1.67 & 1.92 & 1.60 & 1.56 & 1.43\\
 & X-Transfer & 0.25 & 2.29 & 0.67 & 2.00 & 1.71 & 1.80 & 2.00 & 2.29 & 1.44 & 1.83 & 2.00 & 1.61 & 1.14\\ 
 & \textbf{CAVALRY-V} & 0.25 & 1.71 & 0.67 & 2.09 & 1.50 & 1.40 & 2.00 & 2.29 & 1.56 & 2.08 & 1.40 & 1.56 & 0.71\\
 \shline
\end{tabular}
\caption{Hal: Hallucination, SIT: Structuralized Image-Text Understanding, MC: Mathematical Calculation, PP: Physical Property, FR: Function Reasoning, IR: Identity Reasoning, NR: Natural Relation, PR: Physical Relation, SR: Social Relation, CSR: Common Sense Reasoning, CFR: Counterfactual Reasoning, CR: Causal Reasoning, FP: Future Prediction.}
\label{tab:main-results-fine-2}
\end{table*}

\section{Additional Experimental Results}

\paragraph{Fine-grained Performance Analysis}

Tables \ref{tab:main-results-fine-1} and \ref{tab:main-results-fine-2} present the detailed category-specific results that constitute the overall scores reported in Table \ref{tab:video_results}. These supplementary tables provide a fine-grained breakdown across all evaluation dimensions.
Table \ref{tab:main-results-fine-1} details performance across perception-related tasks, including Video Topic (VT), Video Emotion (VE), Video Scene (VS), Video Style (VSt), OCR capabilities, Object Recognition (OR), Attribute Recognition (AR), Event Recognition (ER), Human Motion (HM), Counting (C), Spatial Relationship (SR), Human-object Interaction (HOI), and Human Interaction (HI). Consistent with our main findings, CAVALRY-V demonstrates substantial improvements in visually-grounded perception tasks.
Table \ref{tab:main-results-fine-2} extends this analysis to reasoning capabilities, covering Hallucination (Hal), Structuralized Image-Text Understanding (SIT), Mathematical Calculation (MC), Physical Property (PP), Function Reasoning (FR), Identity Reasoning (IR), Natural Relation (NR), Physical Relation (PR), Social Relation (SR), Common Sense Reasoning (CSR), Counterfactual Reasoning (CFR), Causal Reasoning (CR), and Future Prediction (FP). The overall scores reported in Table \ref{tab:video_results} are computed as weighted averages of these fine-grained metrics, providing a comprehensive assessment of attack effectiveness while maintaining consistency with established benchmark practices.

\end{document}